\documentclass[runningheads]{llncs}

\usepackage{amsmath}
\usepackage{comment}
\usepackage{multirow}
\usepackage{makecell}
\usepackage{url}
\usepackage{xcolor}
\usepackage{ctable}
\usepackage{subcaption}
\usepackage{hyperref}
\hypersetup{
    colorlinks=true,
    linkcolor=blue,
    filecolor=blue,
    urlcolor=blue,
    citecolor=blue
    }

\usepackage[T1]{fontenc}
\usepackage{graphicx}
\begin{document}
\title{Interpretable Medical Image Classification using Prototype Learning and Privileged Information}
\titlerunning{Proto-Caps: Interpretable Medical Image Classification}

\author{Luisa Gallée\inst{1}\orcidID{0000-0001-5556-7395} \and Meinrad Beer \inst{2}\orcidID{0000-0001-7523-1979} \and
Michael Götz\inst{1,3}\orcidID{0000-0003-0984-224X}}

\authorrunning{Gallée et al.}

\institute{Experimental Radiology, University Hospital Ulm, Germany \email{\{luisa.gallee,michael.goetz\}@uni-ulm.de} \and  Department of Diagnostic and Interventional Radiology, University Hospital Ulm, Germany  \and i2SouI - Innovative Imaging in Surgical Oncology Ulm, University Hospital Ulm, Germany}

\maketitle 
\begin{abstract}
Interpretability is often an essential requirement in medical imaging.
Advanced deep learning methods are required to address this need for explainability and high performance. In this work, we investigate whether additional information available during the training process can be used to create an understandable and powerful model. We propose an innovative solution called \textit{Proto-Caps} that leverages the benefits of capsule networks, prototype learning and the use of privileged information.
Evaluating the proposed solution on the LIDC-IDRI dataset shows that it combines increased interpretability with above state-of-the-art prediction performance. Compared to the explainable baseline model, our method achieves more than 6\,$\%$ higher accuracy in predicting both malignancy (93.0\,$\%$) and mean characteristic features of lung nodules. Simultaneously, the model provides case-based reasoning with prototype representations that allow visual validation of radiologist-defined attributes.

\keywords{Explainable AI  \and Capsule Network \and Prototype Learning}
\end{abstract}
\section{Introduction}

Deep learning-based systems show remarkable predictive performance in many computer vision tasks, including medical image analysis, and are often comparable to human performance.
However, the complexity of this technique makes it challenging to extract model knowledge and understand model decisions. 
This limitation is being addressed by the field of Explainable AI, in which significant progress has been made in recent years. 
An important line of research is the use of inherently explainable models, which circumvent the need for indirect, error-prone on-top explanations \cite{rudin2019stop}. A common misconception is that the additional explanation comes with a decrease in performance. However, Rudin et al. \cite{rudin2019stop} and others have already pointed out that this can be avoided by designing algorithms that build explainability into the core concept, rather than just adding it on top. Our work proves this once again by providing a powerful and explainable solution for medical image classification.

A promising approach for interpretability is the use of \textbf{Privileged Information}, i.e. information that is only available during training \cite{vapnik2015learning,VAPNIK2009544}. Besides using the additional knowledge to improve performance, it can also help to increase explainability, as has already been shown using the LIDC-IDRI dataset \cite{armato_iii_lung_data_2015}. In addition to the malignancy of the lung nodules, which is the main goal of the prediction task, the radiologists also marked certain nodule characteristics such as sphericity, margin or spiculation. Shen et al. \cite{shen_interpretable_2019} used the attributes with a hierarchical 3D CNN approach, demonstrating the potential of using this privileged information. LaLonde et al. \cite{lalonde_encoding_2020} extended this idea using capsule networks, a technique for learning individual, encapsulated representations rather than general convolutional layers \cite{afshar_brain_2018,sabour_dynamic_2017}. This method was used to jointly learn the predefined attributes in the capsules and their associations with the classification target, i.e. malignancy.
Explainability is enabled by providing additional attribute values that are essential to the model output. However, the predicted, possibly incorrect scores for the attributes must be trusted, which raises the question of whether there is a way to validate the predictions.

\textbf{Prototype Networks} are another line of research implementing the idea that the representations of images cluster around a prototypical representation for each class \cite{snell_prototypical_2017}. The goal is to find embedded prototypes (i.e. examples) that best separate the images by their classes \cite{bien_prototype_2011}. This idea has been applied to various methods, such as unsupervised learning \cite{pan_transferrable_2019}, few- and zero-shot learning \cite{snell_prototypical_2017,sun_hierarchical_2019,xu_attribute_2020}, as well as for capsule networks \cite{wang_dynamic_2022}, however without the use of privileged information. A successful approach is prototypical models with case-based reasoning, which justify their prediction by showing prototypical training examples similar to the input instance \cite{barnett_case-based_2021,li_deep_2018}. This idea can be used for region-wise prototypical samples \cite{chen_this_2019}. However, these networks can only tell which prototypical samples resemble the query image, not why. Similar to attention models, regional explanations are learned and provided \cite{zheng_learning_2017,zhou_learning_2016}. It is up to the user to guess which features of the image regions are relevant to the network and are exemplified by the prototypes. 

Our method addresses the limitations of privileged information-based and prototype-based explanation  by combining case-based visual reasoning through exemplary representation of high-level attributes to achieve explainability and high-performance.
The proposed method is an image classifier that satisfies explainable-by-design with two elements: First, decisive intermediate results of a high-performance CNN are trained on human-defined attributes which are being predicted during application.
Second, the model provides prototypical natural images to validate the attribute prediction.
In addition to the enhanced explainability offered by the proposed approach, to our knowledge the proposed method outperforms existing studies on the LIDC-IDRI dataset.

The main contributions of our work are:
\begin{itemize}
\item A novel method that, for the first time to our knowledge, combines privileged information and prototype learning to provide increased explanatory power for medical classification tasks.
\item A prototype network architecture based on a capsule network that leverages the benefits of both techniques.
\item An explainable solution outperforming state-of-the-art explainable and non-explainable methods on the LIDC-IDRI dataset.
\end{itemize}
We provide the code with the model architecture and training algorithm of \textit{Proto-Caps} on GitHub.

\section{Methods}
\begin{figure}[h]
\centering
\includegraphics[width=1\textwidth]{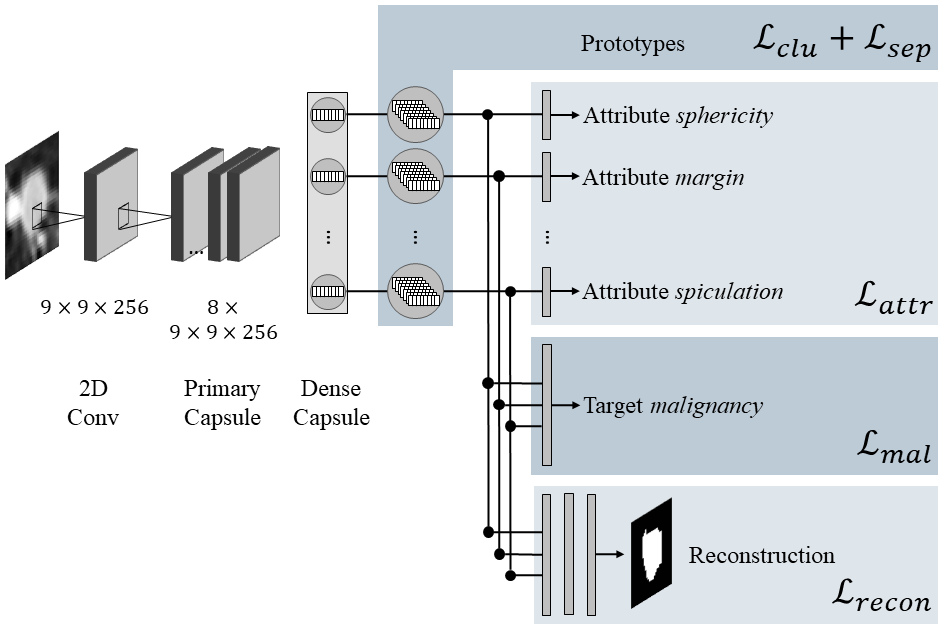}
\caption{\textbf{Proposed Model Architecture} The backbone capsule network results in capsules representing predefined attributes. For each capsule, a set of prototypes is trained. To fit the attribute scores, the capsule vectors are fed through individual dense layers. The latent vectors of all capsules are being accumulated for a dense layer to predict a target score and for a decoder network to reconstruct the region of interest.} \label{fig:model}
\end{figure}

The idea behind our approach is to combine the potential of attribute and prototype learning for a powerful and interpretable learning system. 
For this, we use a capsule network of attribute capsules from which the target class is predicted.
As the attribute prediction can also be susceptible to error, we use prototypes to explain the predictions made for each attribute. Based on \cite{lalonde_encoding_2020}, our approach, called \textit{Proto-Caps}, consists of a backbone capsule network. The network is trained using multiple heads. An attribute head is used to ensure that each capsule represents a single attribute, a reconstruction head learns the original segmentation, and the main target prediction head learns the final classification. 
The model is extended by a prototype layer that provides explanations for each attribute decision.
The overall architecture of \textit{Proto-Caps} is shown in Figure \ref{fig:model}.

The \textbf{backbone} of our approach is a capsule network consisting of three layers: Features of the input image of size $1\times32\times32$ are extracted by a 2D convolutional layer containing 256 kernels of size $9\times9$. 
We decided not to use 3D convolutional layers, as preliminary experiments showed only marginal differences (within std. dev. of results), but required significantly more computing time.
The primary capsule layer then segregates low-level features into 8 different capsules, with each capsule applying 256 kernels of size $9\times9$. The final dense capsule layer consists of one capsule for each attribute and extracts high-level features, overall producing eight 16-dimensional vectors. These vectors form the starting point for the different prediction branches. 

The \textbf{target head}, a fully connected layer, combines the capsule encodings. The loss function for the malignancy prediction was chosen according to LaLonde et al. \cite{lalonde_encoding_2020}, where the distribution of radiologist malignancy annotations is optimized with the Kullback-Leibler divergence $\mathcal{L}_{mal}$ to reflect the inter-observer agreement and thus uncertainty. The \textbf{reconstruction branch} to predict the segmentation mask of the nodule consists of a simple decoder with three fully connected layers with the output filters 512, 1024, and the size of the resulting image $1\times32\times32$. The reconstruction loss $\mathcal{L}_{recon}$ implements the mean square error between the output and the binary segmentation mask. It has been shown that incorporating reconstruction learning is beneficial to performance \cite{lalonde_encoding_2020}. 

For the \textbf{attribute head}, we propose to use fully connected layers, instead of determining the attribute manifestation by the length of the capsule encoding, as was done previously \cite{lalonde_encoding_2020}. Each capsule vector is processed by a separate linear layer to fit the respective attribute score. We formulate the attribute loss as
\begin{equation}
\mathcal{L}_{attr} = (1-b)\sum_a \left\Vert Y_a - O_a\right\Vert^2 ,
\end{equation}
where $Y_a$ is the ground truth mean attribute score by the radiologists, $O_a$ is the network score prediction for the $a$-th attribute, and $b$ is a random binary mask allowing semi-supervised attribute learning.

Two \textbf{prototypes} are learned per possible attribute class, resulting in 8-12 prototypes per attribute (i.e. capsule). During the training, a combined loss function encourages a training sample to be close to a prototype of the correct attribute class and away from prototypes dedicated to others, similar to existing approaches \cite{chen_this_2019}. Randomly initialized, the prototypes are a representative subset of the training dataset for each attribute after the training. For this, a cluster cost reduces the Euclidean distance of a sample's capsule vector $O_a$ to the nearest prototype vector $p_j$ of group $P_{a_s}$ which is dedicated to its correct attribute score.
\begin{equation}
\mathcal{L}_{clu} = \frac{1}{A} \sum^A_a \underset{p_j \in P_{a_s}}{\textrm{min}} \left\Vert O_a-p_j\right\Vert_2 .
\end{equation}
In order to clearly distinguish between different attribute specifications, a separation loss is applied to increase the distance to the capsule prototypes that do not have the correct specification, limited by a maximum distance:
\begin{equation}
    \mathcal{L}_{sep} = \frac{1}{A} \sum^A_a \underset{p_j \notin P_{a_s}}{\textrm{min}} \textrm{max}(0,\textrm{dist}_{max}-\left\Vert O_a-p_j \right\Vert_2) .
\end{equation}

Prototype optimization begins after 100 epochs. In addition to fitting the prototypes with the loss function, each prototype is replaced every 10 epochs by the most similar latent vector of a training sample. 
The original image of the training sample is stored and used for prototype visualization.
During inference, the predicted attribute value is set to the ground truth attribute value of the closest prototype, ignoring the learned dense layers in the attribute head at this stage.

The  overall loss function is the following weighted sum, where $\lambda_{recon} = 0.512$ was chosen according to \cite{lalonde_encoding_2020}, and the prototype weights were chosen empirically:
\begin{equation}
    \mathcal{L} = \mathcal{L}_{mal}+\lambda_{recon}\cdot\mathcal{L}_{recon}+\mathcal{L}_{attr}+0.125\cdot(\mathcal{L}_{clu}+0.1\cdot\mathcal{L}_{sep})
\end{equation}

\section{Experiments}

\textbf{Data}
The proposed approach is evaluated using the publicly available LIDC-IDRI dataset consisting of 1018 clinical thoracic CT scans from patients with Non-Small Cell Lung Cancer (NSCLC) \cite{armato_iii_lung_2011,armato_iii_lung_data_2015}. Each lung nodule with a minimum size of 3\,mm was segmented and annotated with a malignancy score ranging from 1-\textit{highly unlikely} to 5-\textit{highly suspicious} by one to four expert raters. Nodules were also scored according to their characteristics with respect to predefined attributes, namely subtlety (difficulty of detection, 1-\textit{extremely subtle}, 5-\textit{obvious}), internal structure (1-\textit{soft tissue}, 4-\textit{air}), pattern of calcification (1-\textit{popcorn}, 6-\textit{absent}), sphericity (1-\textit{linear}, 5-\textit{round}), margin (1-\textit{poorly defined}, 5-\textit{sharp}), lobulation (1-\textit{no lobulation}, 5-\textit{marked lobulation}), spiculation (1-\textit{no spiculation}, 5-\textit{marked spiculation}), and texture (1-\textit{non-solid}, 5-\textit{solid}). The \texttt{pylidc} framework \cite{hancock_lung_2016} is used to access and process the data. The mean attribute annotation and the mean and standard deviation of the malignancy annotations are calculated. The latter was used to fit a Gaussian distribution, which serves as the ground truth label for optimization. Samples with a mean expert malignancy score of 3-\textit{indeterminate} or annotations  from fewer than three experts were excluded in consistency with the literature \cite{hussein_risk_2017,hussein_tumornet_2017,lalonde_encoding_2020}. 

\textbf{Experiment Designs}
To ensure comparability with previous work \cite{hussein_risk_2017,hussein_tumornet_2017,lalonde_encoding_2020}, the main metric used is Within-1-Accuracy, where a prediction within one score is considered correct. Five-fold stratified cross-validation was performed using 10\,\% of the training data for validation and the best run of three is reported. The algorithm was implemented using the \texttt{PyTorch} framework version 1.13 and \texttt{CUDA} version 11.6. A learning rate of $0.5$ was chosen for the prototype vectors and $0.02$ for the other learnable parameters. The batch size was set to 128 and the optimizer was ADAM \cite{kingma_adam_2014}.  
With a maximum of 1000 epochs, but stopping early if there was no improvement in target accuracy within 100 epochs, the experiments lasted an average of three hours on a GeForce RTX 3090 graphics card.
The code is publicly available at \url{https://github.com/XRad-Ulm/Proto-Caps}.

Besides pure performance, the effect of reduced availability of attribute annotations was investigated. This was done by using attribute information only for a randomly selected fraction of the nodules during the training. 

To investigate the effect of prototypes on the network performance, an ablation study was performed. Three networks were compared: \textbf{Proto-Caps} (proposed) including learning and applying prototypes during inference, \textbf{Proto-Caps}$_{\text{w/o~use}}$ where prototypes are only learned but ignored for inference, and \textbf{Proto-Caps}$_{\text{w/o~learn}}$ using the proposed architecture without any prototypes. 

\section{Results}
\begin{figure}
\centering
\includegraphics[width=1\textwidth]{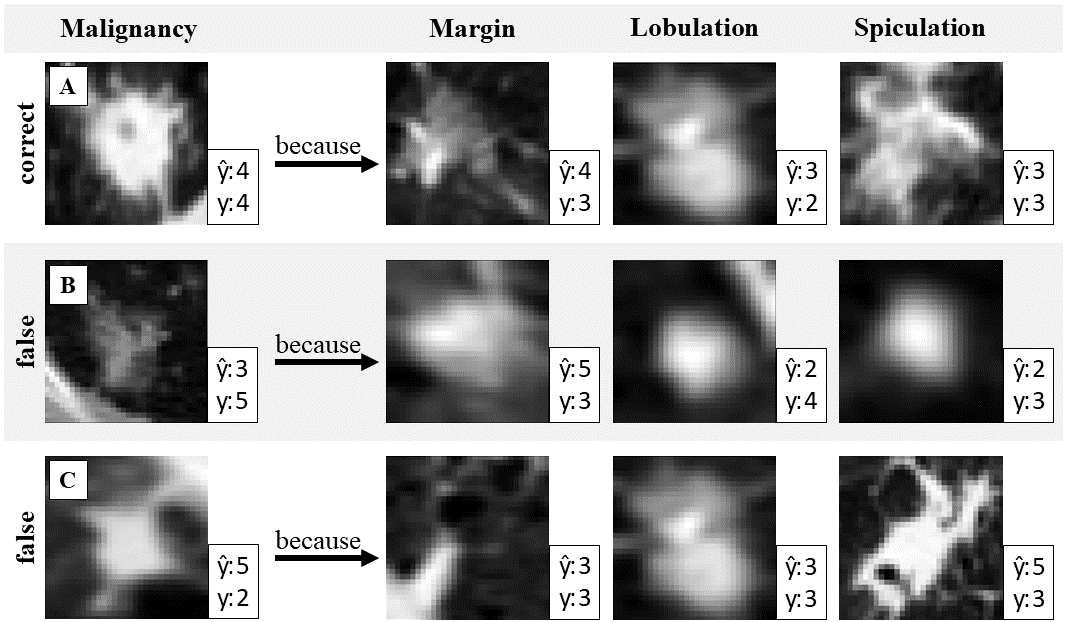}
\caption{One correct and two wrongly predicted examples with exemplary attribute prototypes. Prediction $\hat{\text{y}}$ and ground truth label y of malignancy and attribute respectively.
Identifying false attribute predictions can help to identify misclassification in malignancy.}\label{fig:res_1}
\end{figure}

\textbf{Qualitative}
Figure \ref{fig:res_1} shows examples of model output. The predicted malignancy score is justified by the closest prototypical sample of a certain attribute. The respective original image for each attribute prototype is being saved during the training process and used for visualization during inference.
In case B, there are large differences between the margin and lobulation prototype and the sample. Similarly, in case C, the spiculation prediction is very different from the sample. 

During application, these discrepancies between the prototypes and the sample nodule raise suspicion, and help to assess the malignancy prediction. A quantitative evaluation of the relationship between correctness in attribute and in target prediction using logistic regression analysis shows a strong relationship between both with an accuracy of 0.93/0.1.

\textbf{Quantitative}
Table \ref{tab:results_comparedSOTA} shows the results of our experiments compared to other state-of-the-art approaches, with results taken from original reports. The accuracy of the proposed method exceeds previous work in both the malignancy and almost all attribute predictions, while modelling all given attributes.

\newcolumntype{L}[1]{>{\raggedright\arraybackslash}p{#1}}
\newcolumntype{C}[1]{>{\centering\arraybackslash}p{#1}}
\newcolumntype{R}[1]{>{\raggedleft\arraybackslash}p{#1}}

\begin{table}[h]
\caption{\label{tab:results_comparedSOTA} Comparison with literature values of other works, attribute scores are reported if available. Mean $\mu$ and standard deviation $\sigma$ calculated from 5-fold experiments. Scores reported as Within-1-Accuracy, except for \cite{shen_interpretable_2019}. The best result is in bold.}
\centering
\begin{tabular}{ c c!{\vrule width 2pt}R{0.7cm}|R{0.7cm}|R{0.7cm}|R{0.7cm}|R{0.7cm}|R{0.7cm}|R{0.7cm}|R{0.7cm}|R{0.9cm}  }
 \hline
 &&\multicolumn{8}{c|}{Attribute Prediction Accuracy in \%}& \multirow{2}{*}{\makecell{Malig-\\nancy}}\\
 &&\multicolumn{1}{c|}{Sub}&\multicolumn{1}{c|}{IS}&\multicolumn{1}{c|}{Cal}&\multicolumn{1}{c|}{Sph}&\multicolumn{1}{c|}{Mar}&\multicolumn{1}{c|}{Lob}&\multicolumn{1}{c|}{Spic}&\multicolumn{1}{c|}{Tex}&\\
 \specialrule{2pt}{0pt}{0pt}
 \textbf{Non-explainable}&&&&&&&&&&\\\cline{1-2}
 3D-CNN+MTL \cite{hussein_risk_2017}&&\multicolumn{1}{c|}{-}&\multicolumn{1}{c|}{-}&\multicolumn{1}{c|}{-}&\multicolumn{1}{c|}{-}&\multicolumn{1}{c|}{-}&\multicolumn{1}{c|}{-}&\multicolumn{1}{c|}{-}&\multicolumn{1}{c|}{-}&90.0\\
 TumorNet \cite{hussein_tumornet_2017}&&\multicolumn{1}{c|}{-}&\multicolumn{1}{c|}{-}&\multicolumn{1}{c|}{-}&\multicolumn{1}{c|}{-}&\multicolumn{1}{c|}{-}&\multicolumn{1}{c|}{-}&\multicolumn{1}{c|}{-}&\multicolumn{1}{c|}{-}&92.3\\
 CapsNet \cite{lalonde_encoding_2020}&&\multicolumn{1}{c|}{-}&\multicolumn{1}{c|}{-}&\multicolumn{1}{c|}{-}&\multicolumn{1}{c|}{-}&\multicolumn{1}{c|}{-}&\multicolumn{1}{c|}{-}&\multicolumn{1}{c|}{-}&\multicolumn{1}{c|}{-}&77.0\\
 \hline
\textbf{Explainable}&&&&&&&&&&\\\cline{1-2}
HSCNN \cite{shen_interpretable_2019} (binary ACC.)&&71.9&\multicolumn{1}{c|}{-}&90.8&55.2&72.5&\multicolumn{1}{c|}{-}&\multicolumn{1}{c|}{-}&83.4&84.2\\
 X-Caps \cite{lalonde_encoding_2020}&&\textbf{90.4}&\multicolumn{1}{c|}{-}&\multicolumn{1}{c|}{-}&85.4&84.1&70.7&75.2&93.1&86.4\\
 Proto-Caps (proposed)&$\mu$&89.1&\textbf{99.8}&\textbf{95.4}&\textbf{96.0}&\textbf{88.3}&\textbf{87.9}&\textbf{89.1}&\textbf{93.3}&\textbf{93.0}\\
 &\textcolor{gray}{$\sigma$}&\textcolor{gray}{5.2}&\textcolor{gray}{0.2}&\textcolor{gray}{1.3}&\textcolor{gray}{2.2}&\textcolor{gray}{3.1}&\textcolor{gray}{0.8}&\textcolor{gray}{1.3}&\textcolor{gray}{1.0}&\textcolor{gray}{1.5}\\
 \hline
\end{tabular}
\end{table}

Table \ref{tab:results_ablation} lists the results obtained when only fractions of the training samples come with attribute information. The experiments indicate that the performance of the given approach is maintained up to a fraction of 10\,$\%$.
Using no attribute annotations at all, i.e. no privileged information, achieves a similar performance, but results in a loss of explainability, as the high-level features extracted in the capsules are not understandable to humans.
This result suggests that privileged information here leads to an increase in interpretability for humans by providing attribute predictions and prototypes without interfering with the model performance.

\begin{table}[h]
\caption{\label{tab:results_ablation}Results of data reduction studies where attribute information was only available in fractions of the training dataset.
Mean $\mu$ and standard deviation $\sigma$ calculated from 5-fold experiments. Scores reported as Within-1-Accuracy.}
\centering
\begin{tabular}{ r c!{\vrule width 2pt}R{0.7cm}|R{0.7cm}|R{0.7cm}|R{0.7cm}|R{0.7cm}|R{0.7cm}|R{0.7cm}|R{0.7cm}|R{0.9cm}  }
 \hline
 &&\multicolumn{8}{c|}{Attribute Prediction Accuracy in \%}& \multirow{2}{*}{\makecell{Malig-\\nancy}}\\
 &&\multicolumn{1}{c|}{Sub}&\multicolumn{1}{c|}{IS}&\multicolumn{1}{c|}{Cal}&\multicolumn{1}{c|}{Sph}&\multicolumn{1}{c|}{Mar}&\multicolumn{1}{c|}{Lob}&\multicolumn{1}{c|}{Spic}&\multicolumn{1}{c|}{Tex}&\\
 \specialrule{2pt}{0pt}{0pt}
 100\,\% attribute labels&$\mu$&89.1&99.8&95.4&96.0&88.3&87.9&89.1&93.3&93.0\\
 &\textcolor{gray}{$\sigma$}&\textcolor{gray}{5.2}&\textcolor{gray}{0.2}&\textcolor{gray}{1.3}&\textcolor{gray}{2.2}&\textcolor{gray}{3.1}&\textcolor{gray}{0.8}&\textcolor{gray}{1.3}&\textcolor{gray}{1.0}&\textcolor{gray}{1.5}\\
 10\,$\%$ attribute labels&$\mu$&92.6&99.8&95.7&94.9&90.3&88.8&86.9&92.3&92.4\\
  &\textcolor{gray}{$\sigma$}&\textcolor{gray}{0.9}&\textcolor{gray}{0.2}& \textcolor{gray}{0.9}&\textcolor{gray}{4.1}&\textcolor{gray}{1.6}&\textcolor{gray}{1.6}&\textcolor{gray}{2.4}& \textcolor{gray}{1.4}&\textcolor{gray}{0.8}\\
 1\,$\%$ attribute labels&$\mu$&91.0&99.8&92.8&95.5&79.9&85.7&85.6&91.2&90.2\\
 & \textcolor{gray}{$\sigma$}& \textcolor{gray}{4.5}& \textcolor{gray}{0.2}& \textcolor{gray}{1.4}& \textcolor{gray}{2.3}& \textcolor{gray}{13.1}& \textcolor{gray}{4.4}& \textcolor{gray}{6.8}& \textcolor{gray}{1.7}& \textcolor{gray}{1.1}\\
 0\,$\%$ attribute labels&$\mu$&\multicolumn{1}{c|}{-}&\multicolumn{1}{c|}{-}&\multicolumn{1}{c|}{-}&\multicolumn{1}{c|}{-}&\multicolumn{1}{c|}{-}&\multicolumn{1}{c|}{-}&\multicolumn{1}{c|}{-}&\multicolumn{1}{c|}{-}&92.4\\
 & \textcolor{gray}{$\sigma$}&\multicolumn{1}{c|}{-}&\multicolumn{1}{c|}{-}&\multicolumn{1}{c|}{-}&\multicolumn{1}{c|}{-}&\multicolumn{1}{c|}{-}&\multicolumn{1}{c|}{-}&\multicolumn{1}{c|}{-}&\multicolumn{1}{c|}{-}&\textcolor{gray}{1.0}\\
 \hline
\end{tabular}
\end{table}

The ablation study shows no significant differences between the three models evaluated. 
For the malignancy accuracy, \textbf{Proto-Caps}$_{\text{w/o~use}}$ and \textbf{Proto-Caps}$_{\text{w/o~learn}}$ achieved $\mu=93.9\,\%$ ($\sigma=0.8$) and $\mu=93.7\,\%$ ($\sigma=1.1$), respectively.
The average difference in attribute accuracy compared to the proposed methods is $1.7\,\%$ and $1.5\,\%$ better, respectively, and is more robust across experiments. The best result was obtained when the prototypes were learned but not used, possibly indicating that the prototypes may have a regularising effect during training, but further experiments are needed to confirm this due to the close results. To give an indication of the decoder performance, \textbf{Proto-Caps}$_{\text{w/o~use}}$ achieved a dice score of $79.7\,\%$.

\section{Discussion and Conclusion}

We propose a new method, named \textit{Proto-Caps}, which combines the advantages of privileged information, and prototype learning for an explainable network, achieving more than $6$\,\% better accuracy than the state-of-the-art explainable method. As shown by qualitative results (Figure \ref{fig:res_1}), the obtained prototypes can be used to detect potential false classifications. 
Our method is based on capsule networks, which allow prediction based on attribute-specific prototypes. Compared to class-specific prototypes, our approach is more specific and allows better interpretation of the predictions made. In summary, \textit{Proto-Caps} outputs prediction results for the main classification task and for predefined attributes, and provides visual validation through the prototypical samples of the attributes. The experiments demonstrate that it outperforms state-of-the-art methods that provide less explainability. Our data reduction studies show that the proposed solution is robust to the number of annotated examples, and good results are obtained even with a 90\% reduction in privileged information. This opens the door for application to other datasets by reducing the additional annotation overhead. While we did see a reduction in performance with too few labels, our results suggest that this is mainly due to inhomogeneous coverage of individual attribute values. In this respect, it would be interesting to find out how a specific selection of the annotated samples, e.g. with extremes, affects the accuracies, especially since our results show that the overall performance is robust even when the attributes are not explicitly trained, i.e. without additional privileged information.
Another area of research would be to explore other types of privileged information that require less extra annotation effort, such as medical reports, to train the attribute capsules.
It would also be worth investigating more sophisticated 3D-based capsule networks.

In conclusion, we believe that the approach of leveraging privileged information with comprehensible architectures and prototype learning is promising for various high-risk application domains and offers many opportunities for further research.

\textbf{Acknowledgements} This research was supported by the University of Ulm (Baustein, L.SBN.0214), and the German Federal Ministry of Education and Research (BMBF) within RACOON COMBINE „NUM 2.0“ (FKZ: 01KX2121).
We acknowledge the National Cancer Institute and the Foundation for the National Institutes of Health for the critical role in the creation of the publicly available LIDC/IDRI Dataset.
%
%
%

\begin{thebibliography}{10}
\providecommand{\url}[1]{\texttt{#1}}
\providecommand{\urlprefix}{URL }
\providecommand{\doi}[1]{https://doi.org/#1}

\bibitem{afshar_brain_2018}
Afshar, P., Mohammadi, A., Plataniotis, K.N.: Brain tumor type classification
  via capsule networks. In: 2018 25th {IEEE} international conference on image
  processing ({ICIP}). pp. 3129--3133. IEEE (2018).
  \doi{10.1109/ICIP.2018.8451379}

\bibitem{armato_iii_lung_2011}
Armato~III, S.G., McLennan, G., Bidaut, L., McNitt-Gray, M.F., Meyer, C.R.,
  Reeves, A.P., Zhao, B., Aberle, D.R., Henschke, C.I., Hoffman, E.A.,
  {others}: The lung image database consortium ({LIDC}) and image database
  resource initiative ({IDRI}): a completed reference database of lung nodules
  on {CT} scans. Medical physics  \textbf{38}(2),  915--931 (2011).
  \doi{10.1118/1.3528204}, publisher: Wiley Online Library

\bibitem{armato_iii_lung_data_2015}
Armato~III, S.G., McLennan, G., Bidaut, L., McNitt-Gray, M.F., Meyer, C.R.,
  Reeves, A.P., Zhao, B., Aberle, D.R., Henschke, C.I., Hoffman, E.A.,
  {others}: Data from lidc-idri. The Cancer Imaging Archive  (2015).
  \doi{10.7937/K9/TCIA.2015.LO9QL9SX}, [Data set]

\bibitem{barnett_case-based_2021}
Barnett, A.J., Schwartz, F.R., Tao, C., Chen, C., Ren, Y., Lo, J.Y., Rudin, C.:
  A case-based interpretable deep learning model for classification of mass
  lesions in digital mammography. Nature Machine Intelligence  \textbf{3}(12),
  1061--1070 (Dec 2021). \doi{10.1038/s42256-021-00423-x}

\bibitem{bien_prototype_2011}
Bien, J., Tibshirani, R.: Prototype selection for interpretable classification.
  The Annals of Applied Statistics  \textbf{5}(4),  2403 -- 2424 (2011).
  \doi{10.1214/11-AOAS495}

\bibitem{chen_this_2019}
Chen, C., Li, O., Tao, D., Barnett, A., Rudin, C., Su, J.K.: This looks like
  that: deep learning for interpretable image recognition. Advances in neural
  information processing systems  \textbf{32} (2019)

\bibitem{hancock_lung_2016}
Hancock, M.C., Magnan, J.F.: Lung nodule malignancy classification using only
  radiologist-quantified image features as inputs to statistical learning
  algorithms: probing the {Lung} {Image} {Database} {Consortium} dataset with
  two statistical learning methods. Journal of Medical Imaging  \textbf{3}(4),
  044504--044504 (2016). \doi{10.1117/1.JMI.3.4.044504}

\bibitem{hussein_risk_2017}
Hussein, S., Cao, K., Song, Q., Bagci, U.: Risk stratification of lung nodules
  using {3D} {CNN}-based multi-task learning. In: Information {Processing} in
  {Medical} {Imaging}: 25th {International} {Conference}, {IPMI} 2017, {Boone},
  {NC}, {USA}, {June} 25-30, 2017, {Proceedings} 25. pp. 249--260. Springer
  (2017). \doi{10.1007/978-3-319-59050-9_20}

\bibitem{hussein_tumornet_2017}
Hussein, S., Gillies, R., Cao, K., Song, Q., Bagci, U.: Tumornet: {Lung} nodule
  characterization using multi-view convolutional neural network with gaussian
  process. In: 2017 {IEEE} 14th international symposium on biomedical imaging
  ({ISBI} 2017). pp. 1007--1010. IEEE (2017). \doi{10.1109/ISBI.2017.7950686}

\bibitem{kingma_adam_2014}
Kingma, D.P., Ba, J.: Adam: {A} method for stochastic optimization. arXiv
  preprint arXiv:1412.6980  (2014)

\bibitem{lalonde_encoding_2020}
LaLonde, R., Torigian, D., Bagci, U.: Encoding visual attributes in capsules
  for explainable medical diagnoses. In: International {Conference} on
  {Medical} {Image} {Computing} and {Computer}-{Assisted} {Intervention}. pp.
  294--304. Springer (2020). \doi{10.1007/978-3-030-59710-8_29}

\bibitem{li_deep_2018}
Li, O., Liu, H., Chen, C., Rudin, C.: Deep learning for case-based reasoning
  through prototypes: {A} neural network that explains its predictions. In:
  Proceedings of the {AAAI} {Conference} on {Artificial} {Intelligence}.
  vol.~32 (2018), issue: 1

\bibitem{pan_transferrable_2019}
Pan, Y., Yao, T., Li, Y., Wang, Y., Ngo, C.W., Mei, T.: Transferrable
  prototypical networks for unsupervised domain adaptation. In: Proceedings of
  the {IEEE}/{CVF} conference on computer vision and pattern recognition. pp.
  2239--2247 (2019). \doi{10.1109/CVPR.2019.00234}

\bibitem{rudin2019stop}
Rudin, C.: Stop explaining black box machine learning models for high stakes
  decisions and use interpretable models instead. Nature machine intelligence
  \textbf{1}(5),  206--215 (2019). \doi{10.1038/s42256-019-0048-x}

\bibitem{sabour_dynamic_2017}
Sabour, S., Frosst, N., Hinton, G.E.: Dynamic routing between capsules.
  Advances in neural information processing systems  \textbf{30} (2017)

\bibitem{shen_interpretable_2019}
Shen, S., Han, S.X., Aberle, D.R., Bui, A.A., Hsu, W.: An interpretable deep
  hierarchical semantic convolutional neural network for lung nodule malignancy
  classification. Expert systems with applications  \textbf{128},  84--95
  (2019). \doi{10.1016/j.eswa.2019.01.048}, publisher: Elsevier

\bibitem{snell_prototypical_2017}
Snell, J., Swersky, K., Zemel, R.: Prototypical networks for few-shot learning.
  Advances in neural information processing systems  \textbf{30} (2017)

\bibitem{sun_hierarchical_2019}
Sun, S., Sun, Q., Zhou, K., Lv, T.: Hierarchical attention prototypical
  networks for few-shot text classification. In: Proceedings of the 2019
  conference on empirical methods in natural language processing and the 9th
  international joint conference on natural language processing
  ({EMNLP}-{IJCNLP}). pp. 476--485 (2019). \doi{10.18653/v1/D19-1045}

\bibitem{vapnik2015learning}
Vapnik, V., Izmailov, R., et~al.: Learning using privileged information:
  similarity control and knowledge transfer. J. Mach. Learn. Res.
  \textbf{16}(1),  2023--2049 (2015)

\bibitem{VAPNIK2009544}
Vapnik, V., Vashist, A.: A new learning paradigm: Learning using privileged
  information. Neural Networks  \textbf{22}(5),  544--557 (2009).
  \doi{10.1016/j.neunet.2009.06.042}, advances in Neural Networks Research:
  IJCNN2009

\bibitem{wang_dynamic_2022}
Wang, M., Guo, Z., Li, H.: A dynamic routing {CapsNet} based on increment
  prototype clustering for overcoming catastrophic forgetting. IET Computer
  Vision  \textbf{16}(1),  83--97 (2022). \doi{10.1049/cvi2.12068}, publisher:
  Wiley Online Library

\bibitem{xu_attribute_2020}
Xu, W., Xian, Y., Wang, J., Schiele, B., Akata, Z.: Attribute prototype network
  for zero-shot learning. Advances in Neural Information Processing Systems
  \textbf{33},  21969--21980 (2020)

\bibitem{zheng_learning_2017}
Zheng, H., Fu, J., Mei, T., Luo, J.: Learning multi-attention convolutional
  neural network for fine-grained image recognition. In: Proceedings of the
  {IEEE} international conference on computer vision. pp. 5209--5217 (2017).
  \doi{10.1109/ICCV.2017.557}

\bibitem{zhou_learning_2016}
Zhou, B., Khosla, A., Lapedriza, A., Oliva, A., Torralba, A.: Learning deep
  features for discriminative localization. In: Proceedings of the {IEEE}
  conference on computer vision and pattern recognition. pp. 2921--2929 (2016).
  \doi{10.1109/CVPR.2016.319}

\end{thebibliography}

%
\end{document}